\pdfoutput=1

\documentclass[11pt]{article}

\usepackage[preprint]{acl}

\usepackage{times}
\usepackage{latexsym}
\usepackage{url}
\usepackage{hyperref}
\usepackage[multiple]{footmisc}

\def\xHyphenate#1#2\wholeString {\if#1$%
    \else\transform{#1}%
    \takeTheRest#2\ofTheString\fi}

\def\takeTheRest#1\ofTheString\fi
{\fi \xHyphenate#1\wholeString}

\def\transform#1{\url{#1}\hskip 0pt plus 1pt}

\usepackage{comment}
\usepackage{amsmath,amsthm,amsfonts,amssymb,bm,stmaryrd}
\usepackage{makecell} 
\usepackage{multirow, multicol}
\usepackage{colortbl}
\usepackage{booktabs}
\usepackage{soul}
\usepackage{kotex}
\usepackage{xltabular}
\usepackage{color}

\usepackage[T1]{fontenc}

\usepackage[utf8]{inputenc}

\usepackage{microtype}

\usepackage{inconsolata}

\usepackage{graphicx}
\usepackage{xcolor} 
\usepackage{algorithm}
\usepackage{algpseudocode}

\newcommand{\dhk}[1]{\textcolor{black}{#1}}

\newcommand{\nj}[1]{\textcolor{black}{#1}}

\definecolor{hall_color}{RGB}{239,193,156}
\definecolor{cit_color}{RGB}{146,159,191}
\definecolor{color_wrg_txt}{RGB}{255,0,0}
\definecolor{color_wrg_box}{RGB}{234,153,153}
\definecolor{human_simulating_agent}{RGB}{229,240,219}
\definecolor{merlin_1}{RGB}{236,205,205}
\definecolor{merlin_2}{RGB}{251,231,163}

\title{MERLIN: \underline{M}ultimodal \underline{E}mbedding \underline{R}efinement via \underline{L}LM-based \underline{I}terative \underline{N}avigation for Text-Video Retrieval-Rerank Pipeline}

\renewcommand{\thefootnote}{\fnsymbol{footnote}}
\author{
 \textbf{Donghoon Han{\textsuperscript{1}\footnote[1]{}}},
 \textbf{Eunhwan Park\textsuperscript{1}\footnote[1]{}}, \\
 \textbf{Gisang Lee\textsuperscript{2, 3}\footnote[1]{}},
 \textbf{Adam Lee\textsuperscript{4, 5}},
 \textbf{Nojun Kwak\textsuperscript{6}\footnote[2]{}}
\\
 \textsuperscript{1}Buzzni AI Lab,
 \textsuperscript{2}KAIST,
 \textsuperscript{3}Mathpresso Inc., \\ 
 \textsuperscript{4}UC Berkeley,
 \textsuperscript{5}Fainders AI,
 \textsuperscript{6}Seoul National University
\\
   \texttt{\{owen, jude\}@buzzni.com}, \texttt{bobopack@kaist.ac.kr}, \\ \texttt{alee00@berkeley.edu}, 
   \texttt{nojunk@snu.ac.kr}
}

\begin{document}
\maketitle

\renewcommand{\thefootnote}{\arabic{footnote}}

\begin{abstract}
The rapid expansion of multimedia content has made \nj{it increasingly challenging to retrieve relevant videos from large collections accurately.} 
Recent advancements in text-video retrieval have focused on cross-modal interactions, large-scale foundation model training, and probabilistic modeling, yet often neglect the crucial user perspective, leading to discrepancies between user queries and the content retrieved. To address this, we introduce MERLIN (\underline{\textbf{M}}ultimodal \underline{\textbf{E}}mbedding \underline{\textbf{R}}efinement via \underline{\textbf{L}}LM-based \underline{\textbf{I}}terative \underline{\textbf{N}}avigation), a novel training-free pipeline that leverages Large Language Models (LLMs) for iterative feedback learning. MERLIN refines query embeddings from a user perspective, enhancing alignment between queries and video content through a dynamic question answering process. Experimental results on datasets like MSR-VTT, MSVD, and ActivityNet demonstrate that MERLIN substantially improves R@1, outperforming existing systems and confirming the benefits of integrating LLMs into multimodal retrieval systems for more responsive and context-aware multimedia retrieval\footnote{\url{https://github.com/dhk1349/MERLIN_text_to_video_search.git}}.
\end{abstract}

\section{Introduction}
\begin{figure}[t!]
  \includegraphics[width=\columnwidth]{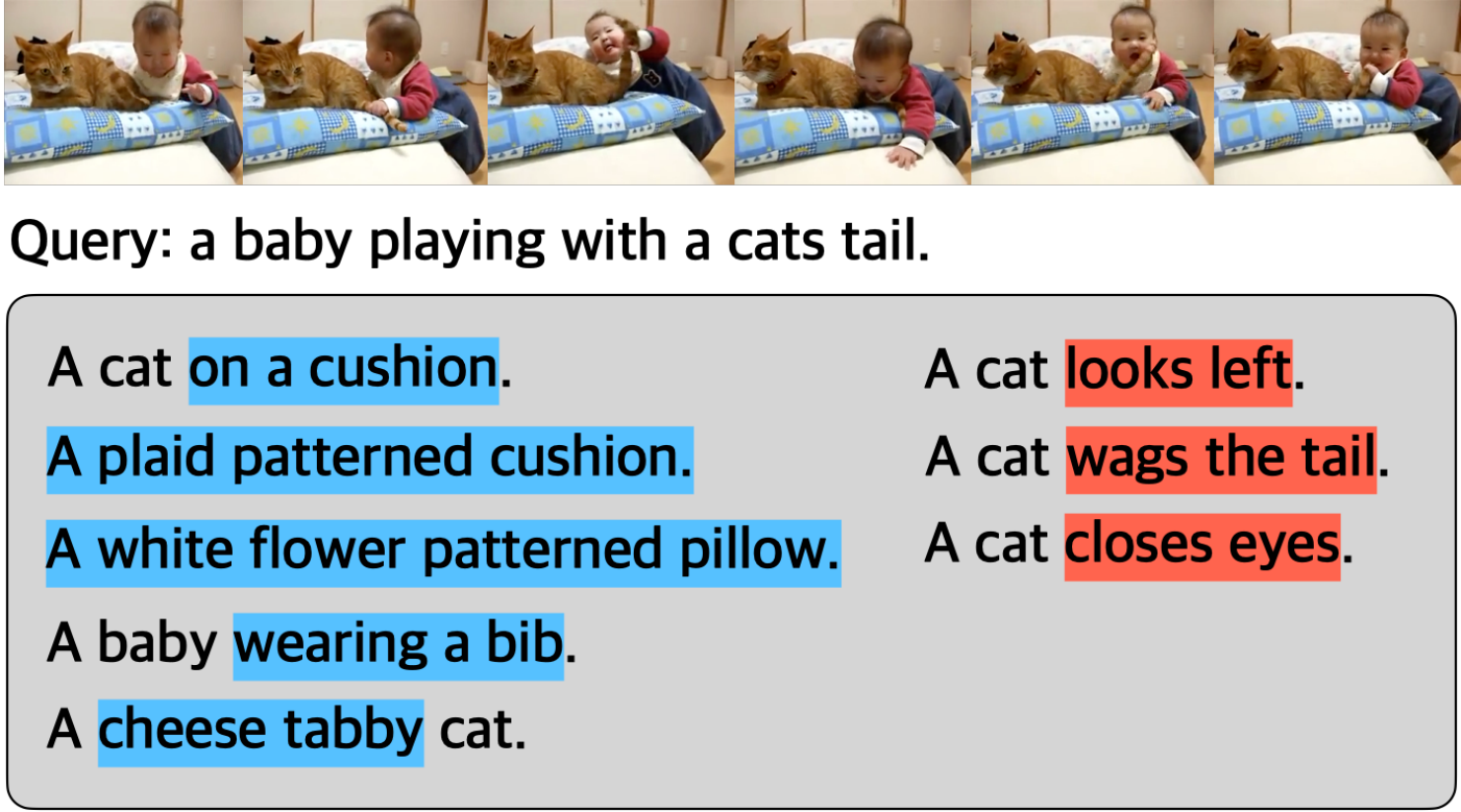}
  \caption{An illustration of the discrepancy between the video caption which could be treated as a user query and the video from MSR-VTT dataset. \textcolor{blue}{Blue} indicates the details that can be observed statically within the video frame, while \textcolor{red}{red} reflects the information that can be obtained temporally across multiple frames.}

  \label{fig:discrepancy_example}
\end{figure}

Multimedia content has recently grown rapidly in both quantity and quality, making the task of finding relevant videos from vast collections increasingly challenging. While recent studies on text-video retrieval have primarily focused on \textit{cross-modal interaction}~\cite{wang2023unified, huang2023vop, wu2023cap4video, jin2023dicosa}, \textit{large-scale foundation model training}~\cite{chen2023panda, chen2023internvl, zhao2024distill, wang2024intern} and \textit{probabilistic modeling}~\cite{hao2023dual, fang2023uatvr, hao2023uncer}, there remains a notable lack of consideration for the discrepancy in text-video retrieval. For instance, as illustrated in Figure~\ref{fig:discrepancy_example}, the video caption \textit{``a baby playing with a cat’s tail''} fails to fully capture the additional context of a playful interaction between the baby and the cat. In real-world scenarios, such discrepancies often arise because users tend to submit succinct queries that do not capture the full context of the videos related to their search intent. Consequently, this mismatch can lead to unsatisfactory retrieval performance. Moreover, neglecting the \textit{user perspective} \nj{makes users} refine their natural language query multiple times to fully reflect their search intent.  \nj{This degrades} the quality of user experience and makes it difficult to understand the search intent, leading to a discrepancy between user queries and the information within the retrieved videos.

To address this issue, we introduce MERLIN (\underline{\textbf{M}}ultimodal \underline{\textbf{E}}mbedding \underline{\textbf{R}}efinement via \underline{\textbf{L}}LM-based \underline{\textbf{I}}terative \underline{\textbf{N}}avigation), a novel training-free and iterative feedback learning pipeline that leverages the power of Large Language Models (LLMs) to augment queries based on the \textit{user perspective}, thereby mitigating the aforementioned discrepancies and significantly improving the text-video retrieval performance. Inspired by human problem-solving and cognitive feedback mechanisms~\cite{flower1981cognitive, DOHERTY1988163}, we employ an interactive and iterative feedback learning~\cite{bohm2019better, Stiennon2020LearningTS, ziegler2019fine, wu2021recursively, ouyang2022training, glaese2022improving, akyurek2023rl4f, madaan2023selfrefine, lee-etal-2024-ask, Liang_2024_CVPR} consisting of a question answering process that iteratively refines query embeddings for text-video retrieval. Moreover, to our best knowledge, MERLIN presents the first implementation of a retrieval–rerank pipeline in the domain of text-video retrieval, establishing a novel framework that prioritizes user intention and interaction in refining search results.

\begin{figure*}[ht!]

  \centering
  \includegraphics[width=0.8\linewidth]{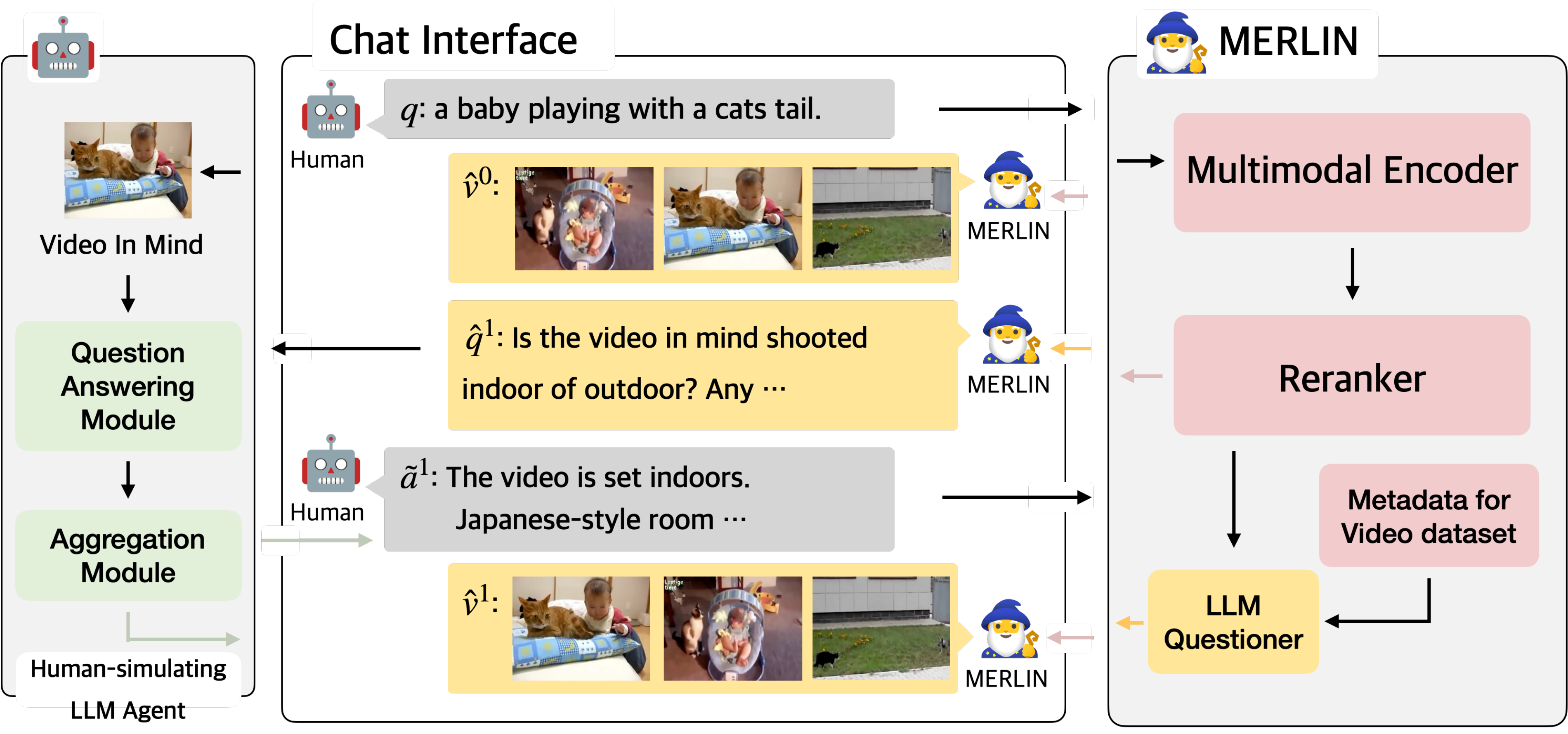}
  \caption {An illustration of MERLIN for text-video retrieval. The \colorbox{merlin_2}{yellow arrow} represents the LLM Questioner returning a question for next round based on metadata of anchor video~(Section~\ref{sec:question_generation}). The \colorbox{human_simulating_agent}{green arrow} represents the human-simulating LLM agent returning an answer based on the ``video in mind'' through Question Answering module and Aggregation module~(Section~\ref{sec:human_simulating_llm_agent}). The \colorbox{merlin_1}{pink arrow} represents MERLIN returning a retrieved video candidates through Multimodal Encoder and Reranker~(Section~\ref{sec:refine}). The system initially retrieves video candidates $\hat{v}^0$ based on the input query text $q$ using a pre-trained multimodal encoder. Using \nj{this} anchor video, LLM Question Generator produces a question $\hat{q}^1$ to elicit additional information from the user~(Section~\ref{sec:question_generation}). The LLM Agent answers this question based on the ``video in mind'', mimicking the human feedback process $\Tilde{a}^1$. The query and answer embeddings are then gradually integrated for each round. The updated query embedding is used to rerank the video candidates $\hat{v}^1$, and the process repeats for multiple rounds. }
  \label{fig:architecture}
\end{figure*}

The primary strength of MERLIN lies in its capability to iteratively adapt and refine query embeddings without necessitating the costly re-training of pre-trained models. 
\dhk{As shown in Figure~\ref{fig:architecture}, when a user submits a query, MERLIN generates questions based on the metadata of the retrieved video candidates and presents these questions to the user. By gathering additional information from the user’s responses, MERLIN refines the embeddings to improve retrieval accuracy, thereby helping users find ``video in mind''\footnote{``video in mind'' refers to the specific video users are looking for or have in mind during the search process.}}.

Experimental results on benchmark datasets, including MSR-VTT, MSVD, and ActivityNet, demonstrate the superiority of the retrieval performance (e.g. R@K) by showing significant improvement. Specifically, MERLIN boosts text-video retrieval performance~(R@1) of Google Multimodal Embedding from $44.00$ to $78.00$ on MSR-VTT, from $52.39$ to $77.61$ on MSVD and from $56.58$ to $68.44$ on ActivityNet.

The key contributions of our paper are as follows: (1) Introduction of MERLIN, a novel LLM-based framework for multimodal embedding refinement that addresses discrepancies between user queries and video content by integrating user perspectives. (2) Implementation of an iterative, cost-effective method for refining query embeddings using LLMs, significantly reducing computational demands while improving retrieval accuracy. (3) Presentation of the first retrieval-rerank pipeline in text-video retrieval, enhancing interactivity and context-awareness within multimodal systems. (4) Experimental results shows that MERLIN substantially improves R@1 on MSR-VTT, MSVD and ActivityNet, thereby demonstrating notable enhancements in zero-shot text-video retrieval.

\section{Related Works}

\paragraph{Dataset.} Text-to-video retrieval aims to retrieve relevant videos based on natural language descriptions and several benchmark video datasets~\cite{anne2017localizing, caba2015activitynet, chen:acl11, xu2016msr} have been curated for this task. One notable dataset is ActivityNet~\cite{caba2015activitynet}, which consists of video-text pairs capturing various human activities. Another widely used dataset is MSR-VTT~\cite{xu2016msr}, which comprises open-domain web videos paired with natural language descriptions. These datasets provide a diverse range of video content and textual queries, enabling comprehensive evaluation of retrieval systems.

\paragraph{Method.} Prior studies have focused on \textit{cross-modal interaction}, \textit{large-scale foundation model training}, and \textit{probabilistic modeling}. In cross-modal interaction~\citet{wang2023unified, huang2023vop, jin2023dicosa} have enhanced reasoning abilities by capturing cross-modal similarities at multiple granularity levels, introduced efficient video prompt mechanisms~\cite{lester-etal-2021-power} with minimal trainable parameters, and improved retrieval with strategies like Disentangled Conceptualization and Set-to-Set Alignment. In foundation model training~\cite{chen2023panda, chen2023internvl, zhao2024distill, wang2024intern}, significant advances have been made with the development of large-scale video and vision-language models leveraging extensive web data, and fine-tuning techniques for better performance on downstream tasks. In probabilistic modeling~\cite{hao2023dual, fang2023uatvr, hao2023uncer}, novel alignment methods and modeling of video and text representations as probabilistic distributions have been proposed to improve text-video retrieval accuracy and address\nj{ed} domain adaptation challenges.

Concurrent to prior studies, \citet{NEURIPS2023_c1b3d1e2} proposed a chat-based image retrieval system (ChatIR) that interacts with users through conversation to gather additional information beyond the initial query, aiming to better understand and clarify the user’s search intent. Following from ChatIR, \cite{lee-etal-2024-interactive} proposed the plug-and-play interactive text-to-image retrieval system. Different from ChatIR, \nj{our} MERLIN incorporates frame-level answer generation tailored to the specific requirements of text-video retrieval, employing a training-free approach. Furthermore, inspired by Composed Image Retrieval~\cite{Liu_2021_ICCV, jang2024spherical}, we iteratively \textit{refine} the embedding by employing spherical linear interpolation, instead \nj{of} iteratively \textit{concatenating} question and answer pair and \nj{feeding} into \nj{the} retrieval model. Lastly, we handle both multi-modality data simultaneously, meaning that our generation result would be more likely aligned to \nj{the} user's search intent. This iterative refinement process mirrors human tendencies to continuously improve their queries based on interactive feedback, akin to strategies seen in feedback-based refinement in textual content. This approach is supported by the growing application of reinforcement learning, which has been increasingly utilized to enhance the quality of generated content through both reference-based and reference-independent feedback mechanisms~\cite{bohm2019better, Stiennon2020LearningTS, ziegler2019fine, wu2021recursively, ouyang2022training, glaese2022improving, akyurek2023rl4f, madaan2023selfrefine, lee-etal-2024-ask, Liang_2024_CVPR}.

\section{Multimodal Embedding Refinement via LLM based Iterative Navigation}

\subsection{Background}
\label{sec_background}

\begin{algorithm}
\small
\caption{Iterative video reranking with question answering rounds}
\begin{algorithmic}[1]
\Require encoder $f_{\text{enc}}()$, user query $q \in \mathcal{Q}$, video $v \in \mathcal{V}$, total question answer round $R$, retrieved top-$k$ videos at round $r$ $\hat{v}^r$, $i$-th candidate among top-$k$ videos at round $r$ $\hat{v}^r_i$, $v^{m}$ a video that user is looking for
\State Encode $\mathbf{e}_q = f_{\text{enc}}(q)$ given user query $q$ 
\State Encode $\mathbf{e}_v = f_{\text{enc}}(v)$ given video $v$ 

\State Retrieve $\hat{v}^0 = \textsc{top-k}_{v\in\mathcal{V}} \big(\textsc{sim} (\mathbf{e}_q, \mathbf{e}_v)\big)$ (Equation~\ref{eq:topk})

\State Initialize message list $m = [ ]$
\For{$r = 1$ to $R$}
    \State Append \text{metadata} of $\hat{v}^{r-1}_0$ to $m$
    \State Generate question $\hat{q}^r=\mathcal{M}_{\mathsf{question}}(m)$~(Equation~\ref{eq:qg})
    \State Append $\hat{q}^r$ to $m$
    \State Generate frame-level answers $\big[\hat{a}^{(r, 0)}, \dots, \hat{a}^{(r, N)}\big] = \mathcal{M}_{\mathsf{answer}}(\hat{q}^r:v^{m})$~(Equation~\ref{eq:vqa})
    \State Aggregate frame-level answers $\Tilde{a}^r = \mathcal{M}_{\mathsf{aggr}}([\hat{a}^{(r, 0)}, \dots, \hat{a}^{(r, N)}\big])$~(Equation~\ref{eq:aggr})
    \State Encode $\mathbf{e}_{A^r} = f_{\text{enc}}(A^r)$
    \State Refine embedding $e = \textsc{Refine}(e_q, \cdots, e_{A^r})$~(Equation~\ref{eq:slerp})
    \State Retrieve $\hat{v}^r = \textsc{top-k}_{v\in\mathcal{V}} \big(\textsc{sim} (\mathbf{e}, \mathbf{e}_v)\big)$ (Equation~\ref{eq:topk})
    
\EndFor
\State \Return Reranked retrieved videos $\hat{v}^r_k$
\end{algorithmic}
\end{algorithm}

Suppose that we have the query text $q \in \mathcal{Q}$, a video $v \in \mathcal{V}$, where $\mathcal{Q}$ and $\mathcal{V}$ indicate a set of queries and videos. Using a pre-trained multimodal encoder $f_{\text{enc}}$, we obtain the query and video embeddings ($\mathbf{e}_q, \mathbf{e}_v)$ as follows:
\begin{align}
    \mathbf{e}_q &= f_{\text{enc}}(q) \in \mathbb{R}^{d}\nonumber \\
    \mathbf{e}_v &= f_{\text{enc}}(v) \in \mathbb{R}^{d}\nonumber,
\end{align}
where $d$ denotes the dimension of embedding. The goal of text-video retrieval is to search the most relevant videos $\hat{v}$'s from a collection of videos $\mathcal{V}$ given a query text $q$ as follows:
\begin{equation}
\label{eq:topk}
    \big[\hat{v}_0, \dots, \hat{v}_{k-1}\big] = \textsc{top-k}_{v\in\mathcal{V}} \big(\textsc{sim} (\mathbf{e}_q, \mathbf{e}_v)\big),
\end{equation}

where $\textsc{sim}(\cdot)$ is a similarity function (e.g., cosine distance, etc). Additionally, our system utilizes two key components: $\mathcal{M}$ and $\mathcal{T}$. Here, $\mathcal{M}$ represents the LLMs and template function $\mathcal{T}$ applies a pre-defined template to inputs\footnote{Note that $\mathcal{M}$ is used interchangeably to indicate both a Large Language Model~(LLM) and a Large Multimodal Model~(LMM).}$^{, }$\footnote{Here, subscripts have been omitted for simplicity. However, subscripts are employed in the equations for each specific module~(e.g., $\mathcal{M}_{\mathsf{question}}$). In addition, the pre-defined template is presented in Appendix due to the limited space.}. Based on this background, we would like to introduce LLM-based iterative navigation, involving multiple rounds of feedback learning and reranking, leading to better performance and interpretability.

\subsection{Question Generation}
\label{sec:question_generation}

Suppose that we have retrieved candidates $\hat{v}^{r}_{k}$ where $r$ and  $k$ indicate the round and the index of the retrieved top $K$ candidates, respectively. We choose $\hat{v}^{r-1}_0$ as an anchor candidate and generate the question with $\mathcal{M}_{\mathsf{question}}$ as follows\footnote{Note that we use the caption from metadata of $\hat{v}^r_0$ and assume that each video consists of $N$ frames.}:
\begin{equation}
    \hat{q}^r = \mathcal{M}_{\mathsf{question}}\big(\mathcal{T}_{\mathsf{question}}(\hat{v}^{r-1}_0)\big).
\label{eq:qg}
\end{equation}
Intuitively, top-ranked candidate is more likely to align with the user's query. This implies that assessing retrieved candidates with question generated from $\hat{v}^{r-1}_0$ using LLMs would enhance retrieval performance and interpretability. 

\subsection{Human-Simulating Agent}
\label{sec:human_simulating_llm_agent}
\paragraph{Video Question Answering.} Our underlying assumption is mitigating the discrepancy between user queries and the information within the videos would be helpful for the better retrieval performance. 

\dhk{To this end, a human-simulating agent answers the question $\hat{q}^r$ with video in mind $v^{m}$, which consists of N frames sampled per second as follows. In this process, we assume a user searching for a specific video, and create a human-simulating agent to mimic the behavior of that user. The agent generates responses by referencing both the video in mind $v^{m}$~(the video the user is looking for) and the questions generated by MERLIN as following:}
\begin{equation}
    \big[\hat{a}^{(r, 0)}, \dots, \hat{a}^{(r, N)}\big] = \mathcal{M}_{\mathsf{answer}}\big(\mathcal{T}_{\mathsf{answer}}(\hat{q}^r), v^{m}\big)
\label{eq:vqa}
\end{equation}
It is worth noting that in a real-world scenario, $\mathcal{M}_{\mathsf{answer}}$ could be replaced by a human. Additionally, using $N$ frames allows us to efficiently handle the temporal information inherent to video, capturing the dynamic aspects of the content. This approach enhances our ability to provide a more comprehensive understanding and alignment with the user's query.

\paragraph{Aggregation.} The individual generated answers for each frame $\big[\hat{a}^{(r, 0)}, \dots, \hat{a}^{(r, N)}\big]$ are now subsequently fed into an \textit{Aggregation Module} which is designed to summarize the multiple frame-level answers into a coherent and concise response to the original query as follows:
\begin{equation}
    \Tilde{a}^r = \mathcal{M}_{\mathsf{aggr}}\Big( \mathcal{T}_{\mathsf{aggr}} \big( [\hat{a}^{(r, 0)}, \dots, \hat{a}^{(r, N)}] \big)\Big).
\label{eq:aggr}
\end{equation}
It is worth noting that Equation~\ref{eq:vqa} provides answers for each frame, however, the summarized answer should capture the importance of the video content. For instance, if the question is ``Did a cookie appear in the video?'' and individual answers for each frame are $\big[\text{``No''}, \text{``No''}, \text{``Yes''}, \text{``No''}\big]$, the Aggregation Module will summarize and provide the final answer for the video as ``Yes'', since a cookie has appeared in the third frame. This process ensures that the temporal and contextual information from all frames is considered, resulting in a more accurate and relevant response.

\subsection{Iterative Embedding Refinement for Reranking}
\label{sec:refine}
Initially, we obtain the answer embedding $\mathbf{e}^{\Tilde{a}^r}$ using the multimodal encoder $f_{\text{enc}}$ as follows: $\mathbf{e}^{\Tilde{a}^r} = f_{\mathsf{enc}} (\Tilde{a}^r)$. Our objective is to dynamically refine the embedding by combining the information from the current round's answer with the previous round's refined embedding. To this end, in the pursuit of refining embeddings iteratively to enhance retrieval performance, we employ a spherical linear interpolation~(SLERP)~\cite{shoemake1985slerp}, which is particularly appropriated for interpolating between embeddings on the unit sphere, preserving the norm and the geometric properties of the embeddings.

Given the embeddings $\mathbf{e}^{r-1}$ from the previous round and $\mathbf{e}^{\Tilde{a}^r}$, the angle $\theta$ between them is computed as:
\begin{equation}
    \theta = \arccos(\mathbf{e}^{\Tilde{a}^r} \cdot \mathbf{e}^{r-1}).
\end{equation}
Note that the angle is essential for determining the interpolation path. Finally, the refined embedding for the current round $\mathbf{e}^r$ is then calculated as:
\begin{equation}
    \mathbf{e}^{r} = \frac{\sin((1-\alpha)\theta)}{\sin(\theta)}\cdot \mathbf{e}^{\Tilde{a}^r}+\frac{\sin(\alpha \theta)}{\sin(\theta)}\cdot\mathbf{e}^{r-1},
\label{eq:slerp}
\end{equation}

where $\alpha \in [0, 1]$ is a hyperparameter that balances the influence of the current answer embedding and the previous refined embedding. This interpolation not only ensures a smooth transition across embedding spaces but also incorporates both the originality of the current response and the semantic context retained from prior interactions. We assume that the potential risk of the iterative embedding refinement is \textit{query drift}~\cite{10.1145/290941.290995, liron2008querydrift, anna2012pred}, a common phenomenon in information retrieval where the focus inadvertently shifts away from the original query intent due to the inclusion of progressively accumulated details. To mitigate the potential risk, we set the $\alpha=0.8$, prioritizing the query and earlier answer embeddings over the most recent answers. We expect that this simple yet effective strategy would preserve the thematic integrity of the initial query, akin to human conversational patterns where early-mentioned topics typically set the context for the entire conversation.

\begin{table*}[tb!]
\centering
\resizebox{0.99\textwidth}{!}{
   \begin{tabular}{l  c ccc    ccc ccc}
    \toprule
    \multirow{2}{*}{\centering \textbf{Model}} & \multirow{2}{*}{\centering \textbf{Rounds}} & \multicolumn{3}{c}{\centering \textbf{MSR-VTT}} &  \multicolumn{3}{c}{\centering \textbf{MSVD}} & \multicolumn{3}{c}{\centering \textbf{ActivityNet}} \\
    \cmidrule(lr){3-5}
    \cmidrule(lr){6-8}
    \cmidrule(lr){9-11}
     & & \textbf{R@1} & \textbf{R@5} & \textbf{R@10} & \textbf{R@1} & \textbf{R@5} & \textbf{R@10}  & \textbf{R@1} & \textbf{R@5} & \textbf{R@10} \\
    
    \midrule
     VAST~\cite{chen2024vast} &  - & 49.30 & 68.30 & 73.90 & - & - & - & - & - & - \\
     InternVideo2-6B~\cite{wang2024internvideo2} & - & 55.9 & 78.3 & 85.1 &  59.3 & 84.4 & 89.6 & 63.2 & 85.6 & 92.5 \\
     LanguageBind-H~\cite{zhu2023languagebind} & - & 44.8 & 70.0 & 78.7 &  53.9 & 80.4 & 87.8 & 41.0 & 68.4 & 80.8 \\
     VideoPrism-g~\cite{madan2024foundation} &  - & 39.7 & 63.7 & - &  - & - & - & 52.7 & 79.4 & - \\
     Marengo-2.6~\cite{marengo-2.6} & - & 49.35 & 73.47 & - &  - & - & - & 55.36 & 82.55 & - \\
    \midrule
     \multirow{6}{*}{\centering MERLIN} 
     &  0 & 44.40 & 67.60 & 76.20 &  52.39 & 77.16 & 84.78 & 56.58 & 84.77 & 91.73 \\
     &  1 & 56.40 & 80.00 & 87.00 &  61.94 & 85.97 & 91.79 & 59.96 & 89.01 & 93.91 \\
     &  2 & 66.40 & 86.00 & 92.80 &  67.61 & 90.45 & 94.63 & 62.68 & 90.42 & 94.34 \\
     &  3 & 72.60 & 91.80 & 95.60 &  71.79 & 91.79 & 96.87 & 66.05 & 90.97 & 95.54 \\
     &  4 & \underline{76.20} & \underline{93.40} & \textbf{97.00} & \underline{74.78} & \underline{93.28} & \underline{96.87} & \underline{67.14} & \underline{91.08} & \underline{95.54} \\
     &  5 & \textbf{78.00} & \textbf{94.20} & \underline{96.80} & \textbf{77.61} & \textbf{94.48} & \textbf{97.31} & \textbf{68.44} & \textbf{91.95} & \textbf{96.63} \\
    \bottomrule
  \end{tabular}
}

\caption{The performance of zero-shot text-video retrieval on MSR-VTT, MSVD, and ActivityNet.}

\label{tab:zs_retrieval}
\end{table*}

\section{Experimental Results}
\subsection{Setting}

To utilize multimodal encoders and LLMs without needing private GPUs, we use Google Multimodal Embedding API\footnote{\url{https://cloud.google.com/generative-ai-studio}} for encoding video and text, and the OpenAI GPT-4o API~\cite{achiam2023gpt}\footnote{\url{https://chat.openai.com/}} for generating questions and answers. These APIs offers comparable performance and reproducibility on benchmarks without private GPUs.

We evaluate MERLIN across three datasets: MSR-VTT, MSVD, and ActivityNet. For MSR-VTT, we sampled 500 videos from its 1,000-sample validation split. From MSVD and ActivityNet, we sampled all 670 and 919 videos from their respective test sets. For videos with multiple captions, we randomly selected one query per video.

\subsection{Performance on Text-Video Retrieval}
The performance of our system is presented in Table~\ref{tab:zs_retrieval}, demonstrating its efficacy through multiple rounds of feedback learning, reflecting the system's ability to iteratively \nj{refine} and incorporate feedback. Particularly, MERLIN shows significant improvements with each round of feedback: On the MSR-VTT dataset, MERLIN shows improvements \nj{of} R@1 \nj{from} $44.40$ to $78.00$, on the MSVD from $52.39$ to $77.61$, and on ActivityNet from $56.58$ to $68.44$ by the final round.

This highlights MERLIN's capacity to adapt and enhance its response through iterative feedback learning. Despite the distinct challenges posed by each dataset, MERLIN significantly boosts its performance, thereby affirming the effectiveness of leveraging iterative feedback learning to enhance text-video retrieval task.

\subsection{Average Ranking of QA Rounds}

\begin{figure}[ht!]
  \includegraphics[width=1.0\linewidth]{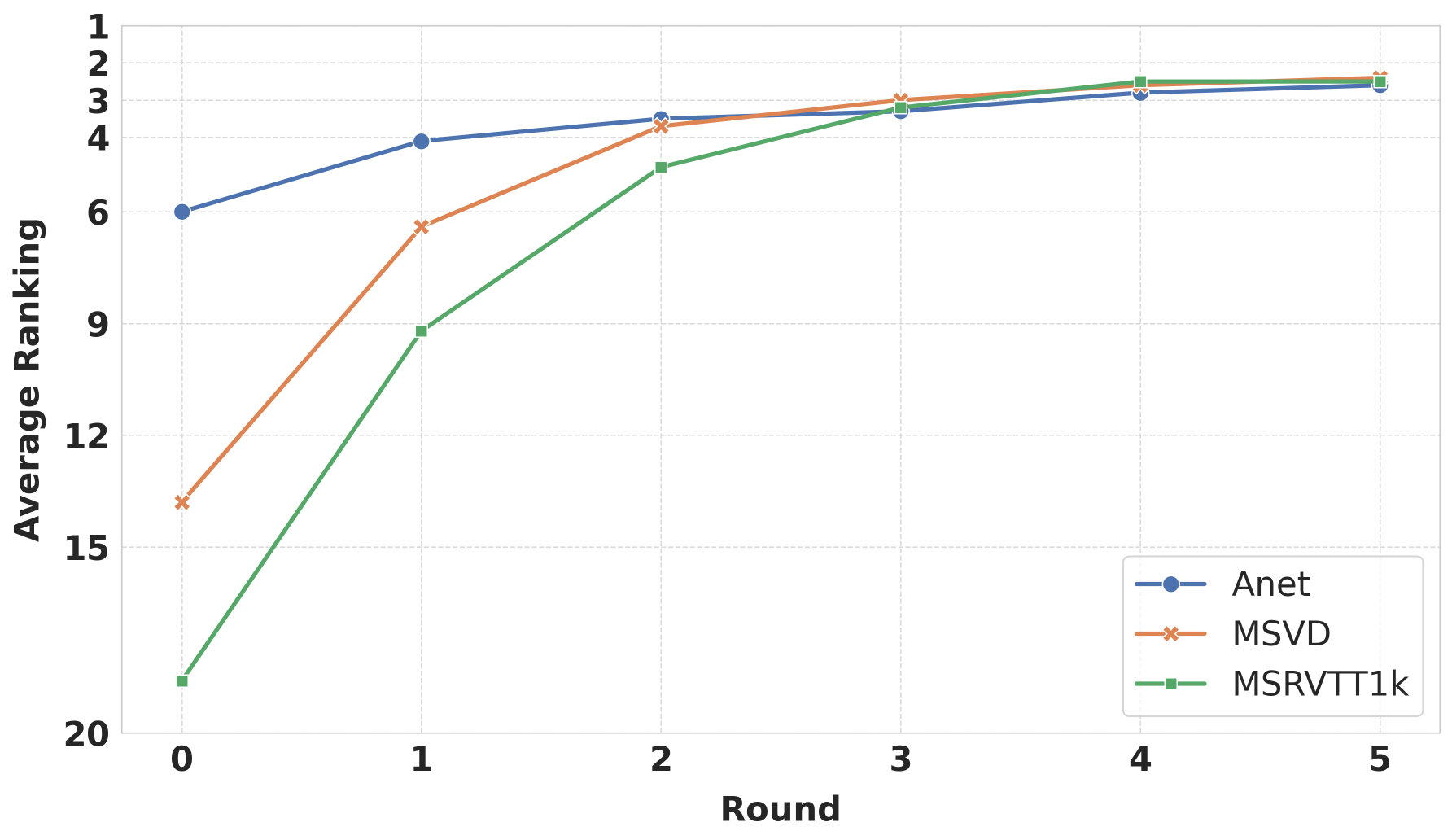}
  \caption{An illustration of the average ranking of target video for each dataset.}
\label{fig:avg_rank}
\end{figure}

In addition to the retrieval performance presented in  Table~\ref{tab:zs_retrieval}, the effectiveness of the iterative query enrichment is further highlighted by examining the average ranking of the target videos across question answer rounds. This analysis is helpful for understanding how the process enhances the ranking of the target videos. As illustrated Figure~\ref{fig:avg_rank}, the average ranking of the target video consistently improves each consecutive round across all datasets. For instance, on the MSR-VTT dataset, the average ranking significantly improves from $18.57$ in round $0$ to $2.5$ by \nj{the} final round. Similar improvements are observed on other datasets, with the average ranking on MSVD improving from $13.84$ to $2.4$, and on ActivityNet from $6$ to $2.6$. This demonstrates the consistent improvement, thereby confirming the effectiveness of MERLIN in reranking through iterative feedback learning.

\section{Ablation Study}
\begin{table}[ht!]
\centering
\resizebox{\linewidth}{!}{
   \begin{tabular}{l  c ccc}
    \toprule
    \multirow{2}{*}{\centering \textbf{Model}} & \multirow{2}{*}{\centering \textbf{Rounds}} & \multicolumn{3}{c}{\centering \textbf{MSR-VTT}}  \\
    \cmidrule(lr){3-5}
     & & \textbf{R@1} & \textbf{R@5} & \textbf{R@10} \\
    \midrule
    Final Query Retrieval~(FQR) & 5 & 51.40 & 71.00 & 78.80 \\
    Refined Reranking~(RR) & 5 & 53.60 & 74.40 & 81.80 \\
    MERLIN & 5 & \textbf{78.00} & \textbf{94.20} & \textbf{96.80} \\ 
    \bottomrule
  \end{tabular}
}
\caption{The performance comparison of video retrieval performance on MSR-VTT using R@K between Final Query Retrieval (FQR), Refined Reranking (RR), and MERLIN. It is worth noting that FQR and RR employ the generated query at final round.}

\label{tab:ablation_1}
\end{table}

\paragraph{The iterative embedding refinement improves retrieval performance.} 

The results in Table~\ref{tab:ablation_1} demonstrate the effectiveness of iterative embedding refinement in improving the retrieval performance. \textbf{Final Query Retrieval~(FQR)}, which direct retrieves videos using the generate query, achieves a R@1 of $51.40$. \textbf{Refined Reranking~(RR)}, which applies reranking to the top-100 initial results, improves performance to $53.60$ at R@1. However, MERLIN, which leverages iterative refinement through multiple rounds of interaction between the query and video embeddings, significantly outperforms both methods, reaching a R@1 of $78.00$, demonstrating the advantage of iterative refinement for aligning query representations with video content. The consistent improvements at R@5 and R@10 further highlight the robustness of MERLIN in video retrieval tasks.

\paragraph{The higher $\alpha$ could mitigate the query drift.}

As mentioned in Section~\ref{sec:refine}, our assumption is mitigating \textit{query drift} would preserve the thematic integrity of the initial query by assigning high $\alpha$ value, prioritizing the query and earlier answer embeddings over the most recent answers.

To validate our assumption in contrast to the experiment's higher $\alpha=0.8$, we conduct additional experiments with assigning a reduced value $\alpha=0.2$, which allows us to observe the impact of shifting emphasis towards the latest answers. The results on the MSR-VTT and MSVD datasets show that setting a lower $\alpha$ initially improves retrieval performance in early rounds but leads to a decline after a few rounds, indicating potential \textit{query drift}. Furthermore, the average ranking of the target video deteriorates in later rounds, suggesting the query representation has deviated from the user's original intent.

Specifically, for MSR-VTT, MERLIN got $44.4$/$67.60$/$76.20$ for R@1/5/10 at round $0$ respectively but ended up with $61.6$/$81.20$/$87.00$ respectively at round $5$. For MSVD, MERLIN got $52.39$/$77.16$/$84.78$ for R@1/5/10 at round $0$ respectively but ended up with $56.87$/$78.51$/$84.63$ respectively at round $5$.

\section{Case Study}
The main objective of MERLIN is to improve the ranking of failure cases where the target video is not among the top-ranked candidates. At the same time, it is important to keep the success case to stay in the top-ranked candidates while MERLIN proceeds \nj{to chat} with \nj{the} user. Retrieving the target video among the top-ranked candidates indicates that MERLIN consistently reflects user intention during the conversation.
To qualitatively verify that MERLIN performs its tasks according to the aforementioned objectives, we reviewed several case studies. We focused on how MERLIN \nj{brings} the rank of failure cases.

\paragraph{Case study for ActivityNet}

As shown in Figure~\ref{fig:casestudy_activitynet}, \nj{the} initial ranking of \nj{the} target video was $224$ using \nj{a} paired query from \nj{the} dataset. However as MERLIN augmented \nj{the} query using \nj{the} user's response, \nj{the} rank boosted to $36 \rightarrow 14 \rightarrow 4 \rightarrow 1$ as \nj{the} round proceeded. During the conversation, MERLIN was able to understand that the user was looking for a video about Christmas themes, featuring two people, and involving gift wrapping. It managed to rank the target video on top with the augmented information.

\paragraph{Case study for MSVD}

As shown in Figure~\ref{fig:casestudy_msvd}, the initial ranking of the target video was $154$ using a paired query from the dataset. However as MERLIN augmented the query using the user's response, the rank boosted to $14 \rightarrow 1 \rightarrow 1 \rightarrow 1$ as the round proceeded. During the conversation, MERLIN was able to understand that the user was looking for a video about the NBA All-Star game, broadcasted on TNT and the scoreboard telling 74:75. It managed to rank the target video on top with the augmented information at \nj{an} early round and managed to keep the top rank during multiple rounds.

\paragraph{Case study for MSR-VTT}

As shown in Figure~\ref{fig:casestudy_msrvtt1ka}, the initial ranking of the target video was $361$ using a paired query from the dataset. However as MERLIN augmented the query using the user's response, the rank boosted to $197 \rightarrow 14 \rightarrow 1 \rightarrow1$ as the round proceeded. During the conversation, MERLIN was able to understand the detailed features and gestures of humans featured on ``video in mind''. It managed to rank the target video on top with the augmented information at an early round and managed to keep the top rank during multiple rounds.

\section{Conclusion}
In conclusion, the MERLIN framework addresses a critical gap in the field of text-video retrieval by integrating the often-overlooked user perspective into the retrieval process. This integration is achieved through a novel, training-free pipeline that utilizes LLMs for iterative feedback learning, allowing for the dynamic refinement of query embeddings based on user interactions. MERLIN not only aligns more closely with user intent but also enhances the overall search experience by reducing discrepancies between user queries and retrieved video content.

The implementation of MERLIN shows a significant advancement in multimedia retrieval, introducing the first retrieval-rerank pipeline in this domain. By incorporating iterative feedback mechanisms inspired by human cognitive processes, MERLIN facilitates a more aligned and context-aware approach to text-video retrieval. Our experimental results demonstrate the effectiveness of this approach, with substantial improvements in retrieval performance observed across MSR-VTT, MSVD, and ActivityNet datasets.

\section*{Limitations}
\label{sec:limitaion}
While our results are promising, we acknowledge that we cannot provide a comprehensive guide for adapting MERLIN to different settings, as we have not extensively explored the impact of changing various components. However, the core principle of integrating user feedback to iteratively refine the query embedding appears to be a robust approach, regardless of pipeline components, the specific domain, or data modality. Future work could investigate the generalization of MERLIN to other multimedia retrieval tasks and explore the optimal configurations for different scenarios.

\nj{Another} limitation of our approach lies in the use of a human-simulating LLM agent for answering questions based on static video frames. While this agent aims to mimic the human feedback process, it lacks the capability to grasp temporal information and attributes that require a high-level understanding of motion and dynamics. Since the LLM agent first generates answers based on static images and then aggregates them, it struggles to capture knowledge about direction, speed, and other temporal aspects present in the videos. 

Moreover, as most pre-trained video encoders also have shortcomings in effectively modeling temporal capabilities~\cite{liu2024tempcompass}, our video encoder may be affected by this limitation as well. This creates a kind of chicken-and-egg problem, where video encoders can benefit from temporal-rich information only when they can understand temporal information effectively. Conversely, even if the video question answering module~(or similar counterparts) can handle temporal-rich information, if the video encoder does not possess the same capability, it may not benefit from this information. This temporal modeling challenge is a prevalent issue that the community needs to address collectively.

\section*{Acknowledgements}
Nojun Kwak was supported by
NRF grant (2021R1A2C3006659) and IITP grants (RS-2022-II220320, RS-2021-II211343), all funded by MSIT of the Korean Government. This work was partially supported by Deep Learning Research Group AttentionX.

\bibliography{custom}

\begin{thebibliography}{43}
\providecommand{\natexlab}[1]{#1}

\bibitem[{Achiam et~al.(2023)Achiam, Adler, Agarwal, Ahmad, Akkaya, Aleman, Almeida, Altenschmidt, Altman, Anadkat et~al.}]{achiam2023gpt}
Josh Achiam, Steven Adler, Sandhini Agarwal, Lama Ahmad, Ilge Akkaya, Florencia~Leoni Aleman, Diogo Almeida, Janko Altenschmidt, Sam Altman, Shyamal Anadkat, et~al. 2023.
\newblock Gpt-4 technical report.
\newblock \emph{arXiv preprint arXiv:2303.08774}.

\bibitem[{Aky{\"u}rek et~al.(2023)Aky{\"u}rek, Aky{\"u}rek, Madaan, Kalyan, Clark, Wijaya, and Tandon}]{akyurek2023rl4f}
Afra~Feyza Aky{\"u}rek, Ekin Aky{\"u}rek, Aman Madaan, Ashwin Kalyan, Peter Clark, Derry Wijaya, and Niket Tandon. 2023.
\newblock Rl4f: Generating natural language feedback with reinforcement learning for repairing model outputs.
\newblock \emph{arXiv preprint arXiv:2305.08844}.

\bibitem[{Anne~Hendricks et~al.(2017)Anne~Hendricks, Wang, Shechtman, Sivic, Darrell, and Russell}]{anne2017localizing}
Lisa Anne~Hendricks, Oliver Wang, Eli Shechtman, Josef Sivic, Trevor Darrell, and Bryan Russell. 2017.
\newblock Localizing moments in video with natural language.
\newblock In \emph{Proceedings of the IEEE international conference on computer vision}, pages 5803--5812.

\bibitem[{B{\"o}hm et~al.(2019)B{\"o}hm, Gao, Meyer, Shapira, Dagan, and Gurevych}]{bohm2019better}
Florian B{\"o}hm, Yang Gao, Christian~M. Meyer, Ori Shapira, Ido Dagan, and Iryna Gurevych. 2019.
\newblock \href {https://doi.org/10.18653/v1/D19-1307} {Better rewards yield better summaries: Learning to summarise without references}.
\newblock In \emph{Proceedings of the 2019 Conference on Empirical Methods in Natural Language Processing and the 9th International Joint Conference on Natural Language Processing (EMNLP-IJCNLP)}, pages 3110--3120, Hong Kong, China. Association for Computational Linguistics.

\bibitem[{Caba~Heilbron et~al.(2015)Caba~Heilbron, Escorcia, Ghanem, and Carlos~Niebles}]{caba2015activitynet}
Fabian Caba~Heilbron, Victor Escorcia, Bernard Ghanem, and Juan Carlos~Niebles. 2015.
\newblock Activitynet: A large-scale video benchmark for human activity understanding.
\newblock In \emph{Proceedings of the ieee conference on computer vision and pattern recognition}, pages 961--970.

\bibitem[{Chen and Dolan(2011)}]{chen:acl11}
David~L. Chen and William~B. Dolan. 2011.
\newblock Collecting highly parallel data for paraphrase evaluation.
\newblock In \emph{Proceedings of the 49th Annual Meeting of the Association for Computational Linguistics (ACL-2011)}, Portland, OR.

\bibitem[{Chen et~al.(2024{\natexlab{a}})Chen, Li, Wang, Zhao, Sun, Zhu, and Liu}]{chen2024vast}
Sihan Chen, Handong Li, Qunbo Wang, Zijia Zhao, Mingzhen Sun, Xinxin Zhu, and Jing Liu. 2024{\natexlab{a}}.
\newblock Vast: A vision-audio-subtitle-text omni-modality foundation model and dataset.
\newblock \emph{Advances in Neural Information Processing Systems}, 36.

\bibitem[{Chen et~al.(2024{\natexlab{b}})Chen, Siarohin, Menapace, Deyneka, Chao, Jeon, Fang, Lee, Ren, Yang, and Tulyakov}]{chen2023panda}
Tsai{-}Shien Chen, Aliaksandr Siarohin, Willi Menapace, Ekaterina Deyneka, Hsiang{-}wei Chao, Byung~Eun Jeon, Yuwei Fang, Hsin{-}Ying Lee, Jian Ren, Ming{-}Hsuan Yang, and Sergey Tulyakov. 2024{\natexlab{b}}.
\newblock \href {https://doi.org/10.48550/ARXIV.2402.19479} {Panda-70m: Captioning 70m videos with multiple cross-modality teachers}.
\newblock \emph{CoRR}, abs/2402.19479.

\bibitem[{Chen et~al.(2023)Chen, Wu, Wang, Su, Chen, Xing, Zhong, Zhang, Zhu, Lu, Li, Luo, Lu, Qiao, and Dai}]{chen2023internvl}
Zhe Chen, Jiannan Wu, Wenhai Wang, Weijie Su, Guo Chen, Sen Xing, Muyan Zhong, Qinglong Zhang, Xizhou Zhu, Lewei Lu, Bin Li, Ping Luo, Tong Lu, Yu~Qiao, and Jifeng Dai. 2023.
\newblock \href {https://doi.org/10.48550/ARXIV.2312.14238} {Internvl: Scaling up vision foundation models and aligning for generic visual-linguistic tasks}.
\newblock \emph{CoRR}, abs/2312.14238.

\bibitem[{Doherty and Balzer(1988)}]{DOHERTY1988163}
Michael~E. Doherty and William~K. Balzer. 1988.
\newblock \href {https://doi.org/10.1016/S0166-4115(08)62173-1} {Chapter 5 cognitive feedback}.
\newblock In Berndt Brehmer and C.R.B. Joyce, editors, \emph{Human Judgment the SJT View}, volume~54 of \emph{Advances in Psychology}, pages 163--197. North-Holland.

\bibitem[{Fang et~al.(2023)Fang, Wu, Liu, Zhou, Song, Wang, Shu, Ji, and Wang}]{fang2023uatvr}
Bo~Fang, Wenhao Wu, Chang Liu, Yu~Zhou, Yuxin Song, Weiping Wang, Xiangbo Shu, Xiangyang Ji, and Jingdong Wang. 2023.
\newblock \href {https://doi.org/10.1109/ICCV51070.2023.01262} {{UATVR:} uncertainty-adaptive text-video retrieval}.
\newblock In \emph{{IEEE/CVF} International Conference on Computer Vision, {ICCV} 2023, Paris, France, October 1-6, 2023}, pages 13677--13687. {IEEE}.

\bibitem[{Flower and Hayes(1981)}]{flower1981cognitive}
Linda Flower and John~R. Hayes. 1981.
\newblock \href {http://www.jstor.org/stable/356600} {A cognitive process theory of writing}.
\newblock \emph{College Composition and Communication}, 32(4):365--387.

\bibitem[{Glaese et~al.(2022)Glaese, McAleese, Tr{\k{e}}bacz, Aslanides, Firoiu, Ewalds, Rauh, Weidinger, Chadwick, Thacker et~al.}]{glaese2022improving}
Amelia Glaese, Nat McAleese, Maja Tr{\k{e}}bacz, John Aslanides, Vlad Firoiu, Timo Ewalds, Maribeth Rauh, Laura Weidinger, Martin Chadwick, Phoebe Thacker, et~al. 2022.
\newblock Improving alignment of dialogue agents via targeted human judgements.
\newblock \emph{arXiv preprint arXiv:2209.14375}.

\bibitem[{Hao and Zhang(2023)}]{hao2023uncer}
Xiaoshuai Hao and Wanqian Zhang. 2023.
\newblock \href {http://papers.nips.cc/paper\_files/paper/2023/hash/78526d7ad4a2532bd91416e948b9644c-Abstract-Conference.html} {Uncertainty-aware alignment network for cross-domain video-text retrieval}.
\newblock In \emph{Advances in Neural Information Processing Systems 36: Annual Conference on Neural Information Processing Systems 2023, NeurIPS 2023, New Orleans, LA, USA, December 10 - 16, 2023}.

\bibitem[{Hao et~al.(2023)Hao, Zhang, Wu, Zhu, and Li}]{hao2023dual}
Xiaoshuai Hao, Wanqian Zhang, Dayan Wu, Fei Zhu, and Bo~Li. 2023.
\newblock \href {https://doi.org/10.1109/CVPR52729.2023.01818} {Dual alignment unsupervised domain adaptation for video-text retrieval}.
\newblock In \emph{{IEEE/CVF} Conference on Computer Vision and Pattern Recognition, {CVPR} 2023, Vancouver, BC, Canada, June 17-24, 2023}, pages 18962--18972. {IEEE}.

\bibitem[{Huang et~al.(2023)Huang, Gong, Pan, Jiang, Lv, Li, and Wang}]{huang2023vop}
Siteng Huang, Biao Gong, Yulin Pan, Jianwen Jiang, Yiliang Lv, Yuyuan Li, and Donglin Wang. 2023.
\newblock \href {https://doi.org/10.1109/CVPR52729.2023.00635} {Vop: Text-video co-operative prompt tuning for cross-modal retrieval}.
\newblock In \emph{{IEEE/CVF} Conference on Computer Vision and Pattern Recognition, {CVPR} 2023, Vancouver, BC, Canada, June 17-24, 2023}, pages 6565--6574. {IEEE}.

\bibitem[{Jang et~al.(2024)Jang, Huynh, Shah, Chen, and Lim}]{jang2024spherical}
Young~Kyun Jang, Dat Huynh, Ashish Shah, Wen-Kai Chen, and Ser-Nam Lim. 2024.
\newblock Spherical linear interpolation and text-anchoring for zero-shot composed image retrieval.
\newblock \emph{arXiv preprint arXiv:2405.00571}.

\bibitem[{Jin et~al.(2023)Jin, Li, Cheng, Huang, Wang, Yuan, Liu, and Chen}]{jin2023dicosa}
Peng Jin, Hao Li, Zesen Cheng, Jinfa Huang, Zhennan Wang, Li~Yuan, Chang Liu, and Jie Chen. 2023.
\newblock \href {https://doi.org/10.24963/IJCAI.2023/104} {Text-video retrieval with disentangled conceptualization and set-to-set alignment}.
\newblock In \emph{Proceedings of the Thirty-Second International Joint Conference on Artificial Intelligence, {IJCAI} 2023, 19th-25th August 2023, Macao, SAR, China}, pages 938--946. ijcai.org.

\bibitem[{Labs(2024)}]{marengo-2.6}
Twelve Labs. 2024.
\newblock \href {https://app.twelvelabs.io/blog/introducing-marengo-2-6} {Introducing marengo-2.6-medium}.

\bibitem[{Lee et~al.(2024{\natexlab{a}})Lee, Park, Lee, and Lim}]{lee-etal-2024-ask}
Dongyub Lee, Eunhwan Park, Hodong Lee, and Heuiseok Lim. 2024{\natexlab{a}}.
\newblock \href {https://aclanthology.org/2024.eacl-long.149} {Ask, assess, and refine: Rectifying factual consistency and hallucination in {LLM}s with metric-guided feedback learning}.
\newblock In \emph{Proceedings of the 18th Conference of the European Chapter of the Association for Computational Linguistics (Volume 1: Long Papers)}, pages 2422--2433, St. Julian{'}s, Malta. Association for Computational Linguistics.

\bibitem[{Lee et~al.(2024{\natexlab{b}})Lee, Yu, Park, Yi, and Yoon}]{lee-etal-2024-interactive}
Saehyung Lee, Sangwon Yu, Junsung Park, Jihun Yi, and Sungroh Yoon. 2024{\natexlab{b}}.
\newblock \href {https://doi.org/10.18653/v1/2024.acl-long.46} {Interactive text-to-image retrieval with large language models: A plug-and-play approach}.
\newblock In \emph{Proceedings of the 62nd Annual Meeting of the Association for Computational Linguistics (Volume 1: Long Papers)}, pages 791--809, Bangkok, Thailand. Association for Computational Linguistics.

\bibitem[{Lester et~al.(2021)Lester, Al-Rfou, and Constant}]{lester-etal-2021-power}
Brian Lester, Rami Al-Rfou, and Noah Constant. 2021.
\newblock \href {https://doi.org/10.18653/v1/2021.emnlp-main.243} {The power of scale for parameter-efficient prompt tuning}.
\newblock In \emph{Proceedings of the 2021 Conference on Empirical Methods in Natural Language Processing}, pages 3045--3059, Online and Punta Cana, Dominican Republic. Association for Computational Linguistics.

\bibitem[{Levy et~al.(2023)Levy, Ben-Ari, Darshan, and Lischinski}]{NEURIPS2023_c1b3d1e2}
Matan Levy, Rami Ben-Ari, Nir Darshan, and Dani Lischinski. 2023.
\newblock \href {https://proceedings.neurips.cc/paper_files/paper/2023/file/c1b3d1e2cf53bb28cabd801bd58b3521-Paper-Conference.pdf} {Chatting makes perfect: Chat-based image retrieval}.
\newblock In \emph{Advances in Neural Information Processing Systems}, volume~36, pages 61437--61449. Curran Associates, Inc.

\bibitem[{Liang et~al.(2024)Liang, He, Li, Li, Klimovskiy, Carolan, Sun, Pont-Tuset, Young, Yang, Ke, Dvijotham, Collins, Luo, Li, Kohlhoff, Ramachandran, and Navalpakkam}]{Liang_2024_CVPR}
Youwei Liang, Junfeng He, Gang Li, Peizhao Li, Arseniy Klimovskiy, Nicholas Carolan, Jiao Sun, Jordi Pont-Tuset, Sarah Young, Feng Yang, Junjie Ke, Krishnamurthy~Dj Dvijotham, Katherine~M. Collins, Yiwen Luo, Yang Li, Kai~J Kohlhoff, Deepak Ramachandran, and Vidhya Navalpakkam. 2024.
\newblock Rich human feedback for text-to-image generation.
\newblock In \emph{Proceedings of the IEEE/CVF Conference on Computer Vision and Pattern Recognition (CVPR)}, pages 19401--19411.

\bibitem[{Liu et~al.(2024)Liu, Li, Liu, Wang, Ren, Li, Chen, Sun, and Hou}]{liu2024tempcompass}
Yuanxin Liu, Shicheng Li, Yi~Liu, Yuxiang Wang, Shuhuai Ren, Lei Li, Sishuo Chen, Xu~Sun, and Lu~Hou. 2024.
\newblock Tempcompass: Do video llms really understand videos?
\newblock \emph{arXiv preprint arXiv:2403.00476}.

\bibitem[{Liu et~al.(2021)Liu, Rodriguez-Opazo, Teney, and Gould}]{Liu_2021_ICCV}
Zheyuan Liu, Cristian Rodriguez-Opazo, Damien Teney, and Stephen Gould. 2021.
\newblock Image retrieval on real-life images with pre-trained vision-and-language models.
\newblock In \emph{Proceedings of the IEEE/CVF International Conference on Computer Vision (ICCV)}, pages 2125--2134.

\bibitem[{Madaan et~al.(2023)Madaan, Tandon, Gupta, Hallinan, Gao, Wiegreffe, Alon, Dziri, Prabhumoye, Yang, Welleck, Majumder, Gupta, Yazdanbakhsh, and Clark}]{madaan2023selfrefine}
Aman Madaan, Niket Tandon, Prakhar Gupta, Skyler Hallinan, Luyu Gao, Sarah Wiegreffe, Uri Alon, Nouha Dziri, Shrimai Prabhumoye, Yiming Yang, Sean Welleck, Bodhisattwa~Prasad Majumder, Shashank Gupta, Amir Yazdanbakhsh, and Peter Clark. 2023.
\newblock \href {https://doi.org/10.48550/arXiv.2303.17651} {Self-refine: Iterative refinement with self-feedback}.
\newblock \emph{CoRR}, abs/2303.17651.

\bibitem[{Madan et~al.(2024)Madan, Moegelmose, Modi, Rawat, and Moeslund}]{madan2024foundation}
Neelu Madan, Andreas Moegelmose, Rajat Modi, Yogesh~S Rawat, and Thomas~B Moeslund. 2024.
\newblock Foundation models for video understanding: A survey.
\newblock \emph{arXiv preprint arXiv:2405.03770}.

\bibitem[{Mitra et~al.(1998)Mitra, Singhal, and Buckley}]{10.1145/290941.290995}
Mandar Mitra, Amit Singhal, and Chris Buckley. 1998.
\newblock \href {https://doi.org/10.1145/290941.290995} {Improving automatic query expansion}.
\newblock In \emph{Proceedings of the 21st Annual International ACM SIGIR Conference on Research and Development in Information Retrieval}, SIGIR '98, page 206–214, New York, NY, USA. Association for Computing Machinery.

\bibitem[{Ouyang et~al.(2022)Ouyang, Wu, Jiang, Almeida, Wainwright, Mishkin, Zhang, Agarwal, Slama, Ray et~al.}]{ouyang2022training}
Long Ouyang, Jeffrey Wu, Xu~Jiang, Diogo Almeida, Carroll Wainwright, Pamela Mishkin, Chong Zhang, Sandhini Agarwal, Katarina Slama, Alex Ray, et~al. 2022.
\newblock Training language models to follow instructions with human feedback.
\newblock \emph{Advances in Neural Information Processing Systems}, 35:27730--27744.

\bibitem[{Shoemake(1985)}]{shoemake1985slerp}
Ken Shoemake. 1985.
\newblock \href {https://doi.org/10.1145/325334.325242} {Animating rotation with quaternion curves}.
\newblock In \emph{Proceedings of the 12th Annual Conference on Computer Graphics and Interactive Techniques, {SIGGRAPH} 1985, San Francisco, California, USA, July 22-26, 1985}, pages 245--254. {ACM}.

\bibitem[{Shtok et~al.(2012)Shtok, Kurland, Carmel, Raiber, and Markovits}]{anna2012pred}
Anna Shtok, Oren Kurland, David Carmel, Fiana Raiber, and Gad Markovits. 2012.
\newblock \href {https://doi.org/10.1145/2180868.2180873} {Predicting query performance by query-drift estimation}.
\newblock \emph{ACM Trans. Inf. Syst.}, 30(2).

\bibitem[{Stiennon et~al.(2020)Stiennon, Ouyang, Wu, Ziegler, Lowe, Voss, Radford, Amodei, and Christiano}]{Stiennon2020LearningTS}
Nisan Stiennon, Long Ouyang, Jeff Wu, Daniel~M. Ziegler, Ryan~J. Lowe, Chelsea Voss, Alec Radford, Dario Amodei, and Paul Christiano. 2020.
\newblock Learning to summarize from human feedback.
\newblock \emph{ArXiv}, abs/2009.01325.

\bibitem[{Wang et~al.(2024{\natexlab{a}})Wang, Li, Li, Yu, He, Chen, Pei, Zheng, Xu, Wang, Shi, Jiang, Li, Zhang, Huang, Qiao, Wang, and Wang}]{wang2024intern}
Yi~Wang, Kunchang Li, Xinhao Li, Jiashuo Yu, Yinan He, Guo Chen, Baoqi Pei, Rongkun Zheng, Jilan Xu, Zun Wang, Yansong Shi, Tianxiang Jiang, Songze Li, Hongjie Zhang, Yifei Huang, Yu~Qiao, Yali Wang, and Limin Wang. 2024{\natexlab{a}}.
\newblock \href {https://doi.org/10.48550/ARXIV.2403.15377} {Internvideo2: Scaling video foundation models for multimodal video understanding}.
\newblock \emph{CoRR}, abs/2403.15377.

\bibitem[{Wang et~al.(2024{\natexlab{b}})Wang, Li, Li, Yu, He, Chen, Pei, Zheng, Xu, Wang et~al.}]{wang2024internvideo2}
Yi~Wang, Kunchang Li, Xinhao Li, Jiashuo Yu, Yinan He, Guo Chen, Baoqi Pei, Rongkun Zheng, Jilan Xu, Zun Wang, et~al. 2024{\natexlab{b}}.
\newblock Internvideo2: Scaling video foundation models for multimodal video understanding.
\newblock \emph{arXiv preprint arXiv:2403.15377}.

\bibitem[{Wang et~al.(2023)Wang, Sung, Cheng, Bertasius, and Bansal}]{wang2023unified}
Ziyang Wang, Yi{-}Lin Sung, Feng Cheng, Gedas Bertasius, and Mohit Bansal. 2023.
\newblock \href {https://doi.org/10.1109/ICCV51070.2023.00264} {Unified coarse-to-fine alignment for video-text retrieval}.
\newblock In \emph{{IEEE/CVF} International Conference on Computer Vision, {ICCV} 2023, Paris, France, October 1-6, 2023}, pages 2804--2815. {IEEE}.

\bibitem[{Wu et~al.(2020)Wu, Ouyang, Ziegler, Stiennon, Lowe, Leike, and Christiano}]{wu2021recursively}
Jeff Wu, Long Ouyang, Daniel~M Ziegler, Nisan Stiennon, Ryan Lowe, Jan Leike, and Paul Christiano. 2020.
\newblock Recursively summarizing books with human feedback.
\newblock In \emph{Advances in Neural Information Processing Systems}.

\bibitem[{Wu et~al.(2023)Wu, Luo, Fang, Wang, and Ouyang}]{wu2023cap4video}
Wenhao Wu, Haipeng Luo, Bo~Fang, Jingdong Wang, and Wanli Ouyang. 2023.
\newblock \href {https://doi.org/10.1109/CVPR52729.2023.01031} {Cap4video: What can auxiliary captions do for text-video retrieval?}
\newblock In \emph{{IEEE/CVF} Conference on Computer Vision and Pattern Recognition, {CVPR} 2023, Vancouver, BC, Canada, June 17-24, 2023}, pages 10704--10713. {IEEE}.

\bibitem[{Xu et~al.(2016)Xu, Mei, Yao, and Rui}]{xu2016msr}
Jun Xu, Tao Mei, Ting Yao, and Yong Rui. 2016.
\newblock Msr-vtt: A large video description dataset for bridging video and language.
\newblock In \emph{Proceedings of the IEEE conference on computer vision and pattern recognition}, pages 5288--5296.

\bibitem[{Zhao et~al.(2024)Zhao, Zhao, Zhou, Wu, Chu, Miao, Schroff, Adam, Liu, Gong, Kr{\"{a}}henb{\"{u}}hl, and Yuan}]{zhao2024distill}
Yue Zhao, Long Zhao, Xingyi Zhou, Jialin Wu, Chun{-}Te Chu, Hui Miao, Florian Schroff, Hartwig Adam, Ting Liu, Boqing Gong, Philipp Kr{\"{a}}henb{\"{u}}hl, and Liangzhe Yuan. 2024.
\newblock \href {https://doi.org/10.48550/ARXIV.2401.06129} {Distilling vision-language models on millions of videos}.
\newblock \emph{CoRR}, abs/2401.06129.

\bibitem[{Zhu et~al.(2023)Zhu, Lin, Ning, Yan, Cui, Wang, Pang, Jiang, Zhang, Li et~al.}]{zhu2023languagebind}
Bin Zhu, Bin Lin, Munan Ning, Yang Yan, Jiaxi Cui, HongFa Wang, Yatian Pang, Wenhao Jiang, Junwu Zhang, Zongwei Li, et~al. 2023.
\newblock Languagebind: Extending video-language pretraining to n-modality by language-based semantic alignment.
\newblock \emph{arXiv preprint arXiv:2310.01852}.

\bibitem[{Ziegler et~al.(2019)Ziegler, Stiennon, Wu, Brown, Radford, Amodei, Christiano, and Irving}]{ziegler2019fine}
Daniel~M Ziegler, Nisan Stiennon, Jeffrey Wu, Tom~B Brown, Alec Radford, Dario Amodei, Paul Christiano, and Geoffrey Irving. 2019.
\newblock Fine-tuning language models from human preferences.
\newblock \emph{arXiv preprint arXiv:1909.08593}.

\bibitem[{Zighelnic and Kurland(2008)}]{liron2008querydrift}
Liron Zighelnic and Oren Kurland. 2008.
\newblock \href {https://doi.org/10.1145/1390334.1390524} {Query-drift prevention for robust query expansion}.
\newblock In \emph{Proceedings of the 31st Annual International ACM SIGIR Conference on Research and Development in Information Retrieval}, SIGIR '08, page 825–826, New York, NY, USA. Association for Computing Machinery.

\end{thebibliography}

\clearpage 

\appendix

\section{Prompt Template}
\label{sec:appendix}
\subsection{Prompt for Question Generation Module}

\nj{To} get useful information from a user, it is critical to ask good questions that could \nj{elicit the} user's intention. As depicted in Table~\ref{tab:merline_questioner}, we set \nj{the} top $1$ ranked video as \nj{the} anchor video and prompted GPT-4o to refer to \nj{the} anchor video's metadata. In our case, we used the video's caption as metadata. However, we believe that questions could be more diverse if we could use other data such as Automatic Speech Recognition~(ASR) captions, the characteristics of the video, and so on. As MERLIN proceeds with the chat with the user~(a user-simulating agent), we stacked previous questions and answers and \nj{encouraged} GPT-4 to generate diverse questions without repeating previous ones.

\subsection{Prompts for Human-Simulating Agent}

As \nj{a} human-simulating agent has two steps for answering \nj{the} question regarding ``video in mind'', we have two different settings for each step. This method lacks in understanding direction, speed, and other temporal knowledge as we \nj{discussed in} Limitation. However, we experimentally showed that our human-simulating agent \nj{helps enrich} information.

\subsubsection{Prompt for Question Answering Module}
As depicted in Table~\ref{tab:instruction_human-simulating agent (qa)}, we sampled frames from a video for every 1 second. Then we asynchronously input the sampled frames and the question from MERLIN. We prompted GPT-4o to answer in detail about facts and not just answer \nj{with} ``Yes'' or ``No''. However, this question answering module is the part that takes up \nj{a} large portion of API cost so \nj{the} video may be \nj{sampled} in \nj{a} wider stride to lower the API cost. 

\subsubsection{Prompt for Aggregation Module}
As depicted in Table~\ref{tab:instruction_human-simulating agent (aggr)}, we aggregate all the answers generated from the question answering module. We prompted GPT-4o to aggregate multiple answers made with multiple \nj{frames at the} question answering module and appended an aggregating example. 

\begin{table*}[h]
    \centering
    \small
    \begin{tabular}{p{14cm}}

\toprule
\textbf{Details about question generation module in MERLIN}\\
\toprule
\textbf{System prompt}\\
You are given a caption about a certain video(anchor video) and a query used to retrieve the anchor video. However, this video may not be the exact video that I am looking for.\\ 
\\
Your role is to ask questions about the video I have in mind to get more information about the video. You have 5 rounds and you can only ask one question at a time.\\
\\
Focus on attributes like the number of people, color, shape etc.\\
\\
\textbf{Initial prompt}\\
This is the caption of the retrieved video. Read the video captions and ask some questions to gain more information to help find out the exact video.
Some videos may not have a caption due to an API error saying sorry I can't provide blah blah.
Captions for video: \{anchor video's caption\}\\
\\
Question: \\
\\
\textbf{Question answering round prompt}\\
answer: \{Aggregated answer from user-simulating Agent\} \\
Based on the answer, here's the caption of the reranked video.\\
caption: \{reranked top1 video caption as anchor caption\} \\
Keep asking.\\
\\
Question: \\
\\
\textbf{Max tokens}\\
- 1500\\
\\
\textbf{Temperature}\\
- 0.75\\
        \bottomrule
    \end{tabular}
\caption{
    The instruction and specification for the question generation module in MERLIN using GPT-4o. After initial retrieval at round~$0$, MERLIN generates a question with an initial prompt using the information of the anchor video's caption. After the user answers the question, MERLIN reranks the and generates a question using a new anchor and question answering round prompt.
    }
    \label{tab:merline_questioner}

\end{table*}

\begin{table*}[!ht]
    \centering
    \small
    \begin{tabular}{p{14cm}}

\toprule
\textbf{Details about human-simulating agent (question answering module)}\\
\toprule

\textbf{System Prompt}\\
You are a helpful assistant that answers the question with details. Don't just answer in yes or no. Provide more details(about facts) about the image that might help the questioner.\\
\\
\textbf{Input format}\\
- text: \{Question from MERLIN\} \\
- image: \{Image encoded in base64 captured from video in mind in 1 second interval.\}\\
\\
\textbf{Max tokens}\\
- 50\\
\\
\textbf{Temperature}\\
- 0.3\\
\\
\textbf{Image sampling rate}\\
- 1 second\\
        \bottomrule
    \end{tabular}
\caption{
    The instruction and specification for video question answering human-simulating agent using GPT-4o~(question answering module).
    }
    \label{tab:instruction_human-simulating agent (qa)}

\end{table*}

\begin{table*}[h]
    \centering
    \small
    \begin{tabular}{p{14cm}}

\toprule
\textbf{Details about the human-simulating agent (aggregation module)}\\
\toprule

\textbf{System Prompt}\\
The VQA model is designed to answer questions based on images. 
To apply it to videos, frames are uniformly extracted from the video over time, and the model provides an answer for each frame to a given question. 
This means that for a single question, there will be multiple answers - one for each extracted frame. 
Your role is to review all of the individual answers and summarize them to provide a final answer to the original question. 
When making the final answer, don't use unnecessary words like `Based on the individual answers provided by the VQA model,'. Just answer the question. \\

For example, if the question is ``Did a cookie appear in the video?'' and the individual answers from the frames are [``No'', ``No'', ``Yes'', ``No''], 
then since a cookie appeared in the 3rd frame, you should summarize and answer the question as ``Yes''.
The length of the aggregated answer should be around 30\textasciitilde35 words.\\
\\
\textbf{Input format}\\
Question: \{Question from MERLIN\}\\
VQA Answer: \{Answers from question answering module\}\\
Aggregated Answer: \\
\\
\textbf{Max tokens}\\
- 100\\
\\
\textbf{Temperature}\\
- 0.5\\
        \bottomrule
    \end{tabular}
\caption{
    The instruction and specification for video question answering human-simulating agent using GPT-4o~(aggregation module).
    }
    \label{tab:instruction_human-simulating agent (aggr)}

\end{table*}

\begin{figure*}[h]

  \centering
  \includegraphics[width=0.9\linewidth]{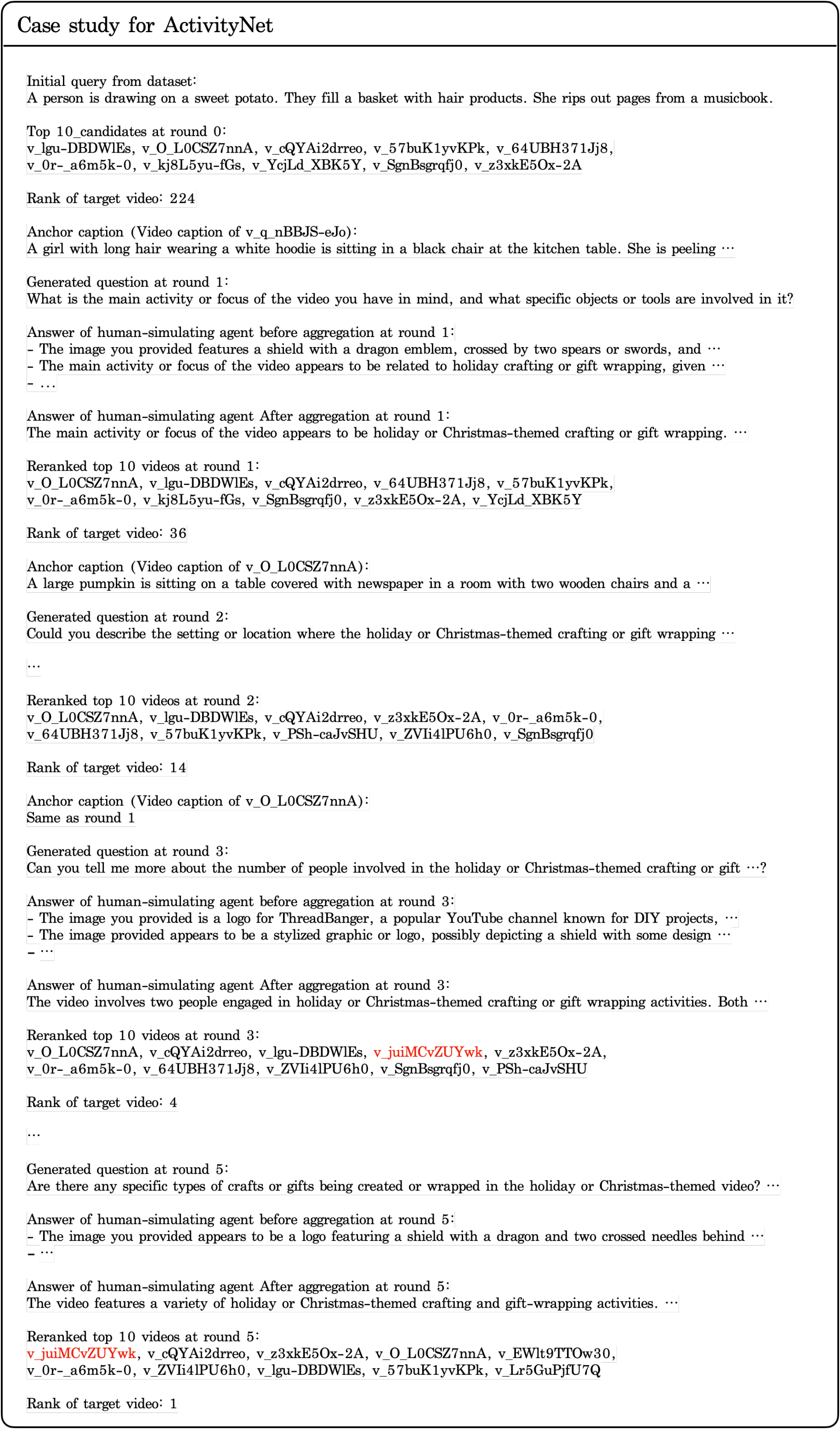}
  \caption {Qualitative evaluation of MERLIN on ActivityNet. sample: v\char`_juiMCvZUYwk.}
  \label{fig:casestudy_activitynet}
\end{figure*}

\begin{figure*}[h]
  \centering
  \includegraphics[width=0.9\linewidth]{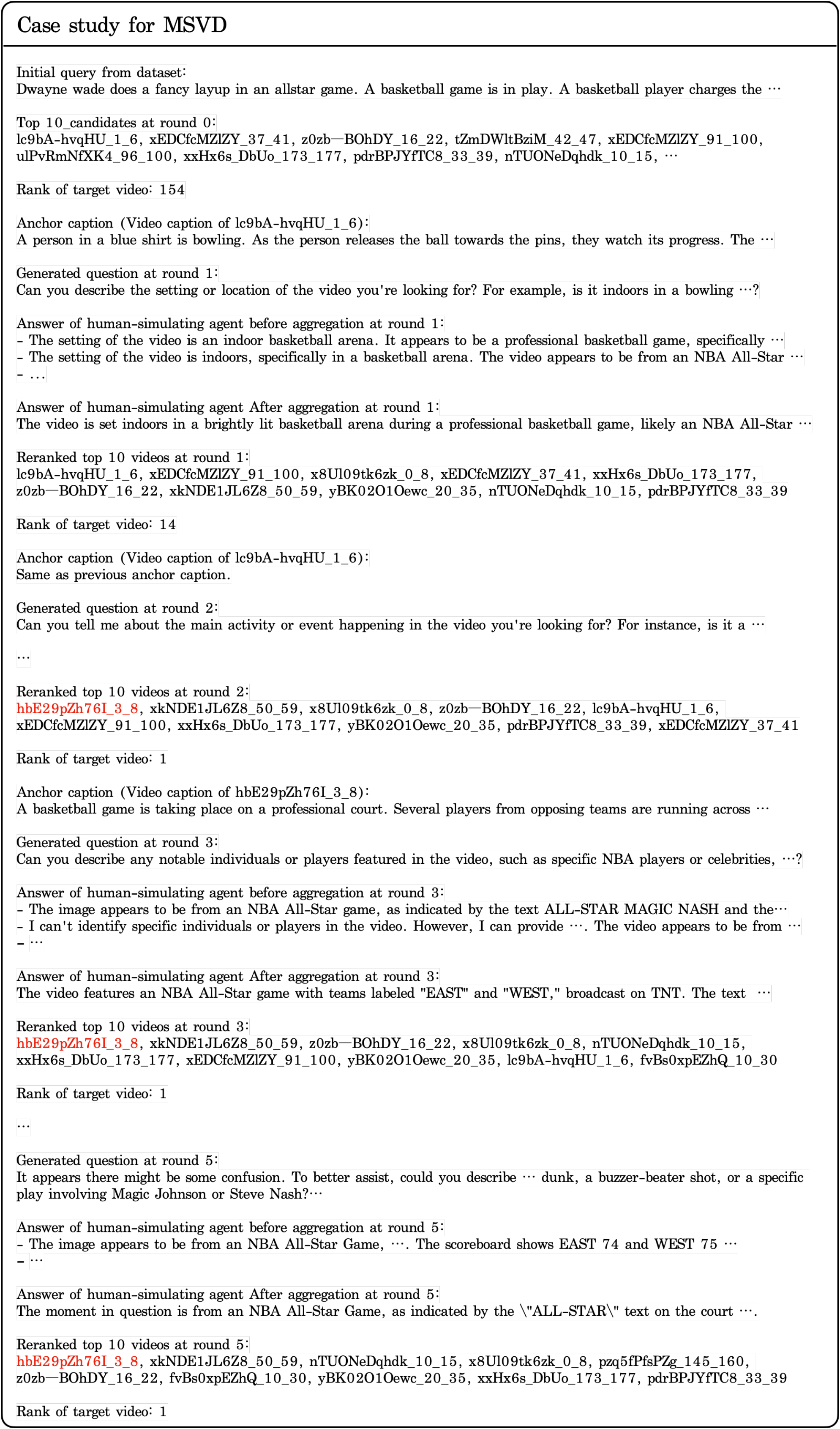}
  \caption {Qualitative evaluation of MERLIN on MSVD. sample:  hbE29pZh76I\char`_3\char`_8.}
  \label{fig:casestudy_msvd}
\end{figure*}

\begin{figure*}[h]
  \centering
  \includegraphics[width=0.9\linewidth]{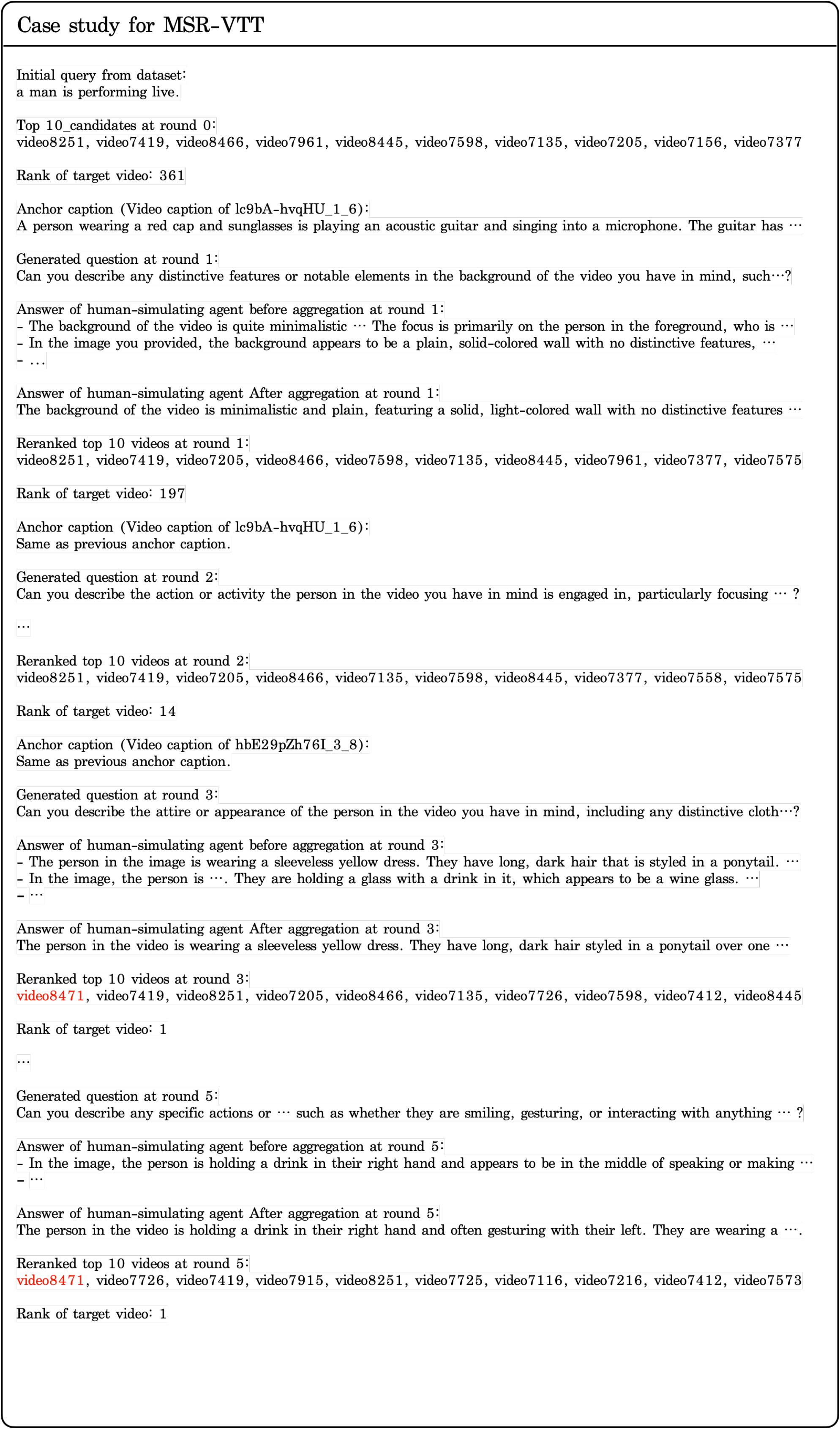}
  \caption {Qualitative evaluation of MERLIN on MSR-VTT1ka. sample: video8471.}
  \label{fig:casestudy_msrvtt1ka}
\end{figure*}

\end{document}